\DeclareMathAlphabet{\mathcal}{OMS}{cmsy}{m}{n}
\DeclareSymbolFont{largesymbols}{OMX}{cmex}{m}{n}

\documentclass{article}
\usepackage{jfrExamplee}

\usepackage{enumitem}
\usepackage{colortbl}
\usepackage{cite} 
\definecolor{mygray}{gray}{0.8}
\usepackage{flushend}
\usepackage{balance}
\usepackage{booktabs}
\usepackage{todonotes}
\usepackage{epsfig}
\usepackage{graphicx}
\usepackage{amsmath}
\usepackage{algorithm}
\usepackage{algorithmic}
\usepackage{amssymb}
\usepackage{lipsum}
\usepackage{subfig}
\usepackage{color}

\usepackage[normalem]{ulem}
\usepackage{bigstrut}
\usepackage{multirow}
\usepackage{siunitx}
\usepackage{ulem}
\usepackage{listings}
\usepackage{eso-pic}
\usepackage{tikz}
\usepackage{xspace}
\usepackage{fancyhdr}
\usepackage{mathtools}
\usepackage{soul}
\usepackage[leftcaption]{sidecap}
\usepackage{hyperref}
\usepackage{listings}
\usepackage{array}
\usepackage{ragged2e}
\usepackage{tcolorbox}
\usepackage{apalike}
\usepackage{setspace}

\PassOptionsToPackage{bookmarks=true,breaklinks=true,letterpaper=true,colorlinks,citecolor=blue,linkcolor=blue,urlcolor=blue}{hyperref}
\usepackage{hyperref}

\title{\LARGE \bf
DaDu-E: Rethinking the Role of Large Language Model in Robotic Computing Pipeline
}

\author{Wenhao Sun$^{1,5,*}$, Sai Hou$^{3,*}$, Zixuan Wang$^{4,5,*}$, Bo Yu$^{2,\dagger}$, Shaoshan Liu$^{2}$,\\
\textbf{Xu Yang$^{3,\dagger}$, Shuai Liang$^{1}$, Yiming Gan$^{1,\dagger}$, Yinhe Han$^{1}$} \\[0.5em]
\small $^{1}$Institute of Computing Technology, Chinese Academy of Sciences (ICT, CAS), Beijing, China \\[-0.3em]
\small $^{2}$Shenzhen Institute of Artificial Intelligence and Robotics for Society (AIRS), Shenzhen, China \\[-0.3em]
\small $^{3}$Beijing Institute of Technology (BIT), Beijing, China \\[-0.3em]
\small $^{4}$Institute of Automation, Chinese Academy of Science (CASIA), Beijing, China \\[-0.3em]
\small $^{5}$University of Chinese Academy of Sciences (UCAS), Beijing, China \\[0.5em]
\small *These authors contributed equally to this work.\\
\small $\dagger$Corresponding authors: Bo Yu, Xu Yang and Yiming Gan (\texttt{ganyiming@ict.ac.cn}).
}

\ifx\figurename\undefined \def\figurename{Figure}\fi
\renewcommand{\figurename}{Fig.}
\renewcommand{\paragraph}[1]{\textbf{#1} }

\newcommand{\Sect}[1]{Sec.~\ref{#1}}

\newcommand{\Tbl}[1]{Tbl.~\ref{#1}}

\newcommand{\no}[1]{#1}
\renewcommand{\no}[1]{}
\newcommand{\RNum}[1]{\uppercase\expandafter{\romannumeral #1\relax}}

\begin{document}

\maketitle
\thispagestyle{empty}
\pagestyle{empty}

\begin{abstract}

Performing complex tasks in open environments remains challenging for robots, even when using large language models (LLMs) as the core planner. Many LLM-based planners are inefficient due to their large number of parameters and prone to inaccuracies because they operate in open-loop systems. We think the reason is that only applying LLMs as planners is insufficient. In this work, we propose DaDu-E, a robust closed-loop planning framework for embodied AI robots. Specifically, DaDu-E is equipped with a relatively lightweight LLM, a set of encapsulated robot skill instructions, a robust feedback system, and memory augmentation. Together, these components enable DaDu-E to (i) actively perceive and adapt to dynamic environments, (ii) optimize computational costs while maintaining high performance, and (iii) recover from execution failures using its memory and feedback mechanisms. Extensive experiments on real-world and simulated tasks show that DaDu-E achieves task success rates comparable to embodied AI robots with larger models as planners like COME-Robot, while reducing computational requirements by $6.6 \times$. Users are encouraged to explore our system at: \url{https://rlc-lab.github.io/dadu-e/}.
\end{abstract}

\begin{center}
\end{center}

\section{INTRODUCTION}
Applying multimodal large language models (LLM) in the planning and decision-making module of the robotic computing stack is becoming a practical solution to improving robots' ability to solve long-horizon tasks. A robot equipped with an LLM as its planner can take flexible instructions directly from users, decomposing a complex task into sequential trivial steps and finishing them to accomplish a complex task. Such an approach significantly improves the usability and performance of traditional program-based robots. 

While prior works mostly focus on integrating LLM into the computing stack and improving the success rate of the tasks~\cite{zhi2024closed,liang2023code,ahn2022can}, they usually refer to the planning ability from the billions of model parameters and countless training data. As a result, these works usually rely on LLMs incompatible running on local servers and are deployed on cloud data centers. Our work pivots from only focusing on performance. Instead of pursuing a higher success rate, we aim to enhance an affordable LLM that can run on a local server with a domain-specific robot skill set, frequent visual feedback, and memory augmentation to achieve a similar success rate while largely reducing the computation cost. 

We propose an efficient LLM planning module for solving long-horizon tasks. Specifically, after being given instructions from the users, the planning module will decompose the tasks into a chain of skills that we will define and program for the robots. After each skill is performed, the planning module will provide one or multiple visual feedback information. The skill chain can be modified based on the visual feedback. On top of the multi-modal LLM, we augment the planning module with a memory module that records recently used objects to save the burden on LLM. 

The design principle of our planning module is twofold. First, we limit the scope of our robots. Specifically, we do not intend to build a general robot planning module capable of doing everything. Instead, we limit the use of the robot to multiple fixed scenarios. For example, in this paper, we limit the planning module to work in the domain of grocery stores and warehouses. With fixed functionality, we can design a lean and efficient skill set for LLMs to use. Second, we enrich the information inputs for the planning module. While most existing works perform open-loop planning and control, we carefully design visual feedback that can be used for closed-loop control and improve the planning results. Moreover, we augment the LLM with a memory unit to store recently used objects and their statuses, which further helps improve efficiency. 

We integrate the above-mentioned planner into a robot computing pipeline. The robot has a moving base and a Universal Robot 03 arm~\cite{UR03}. We evaluate the planner in both simulation environments and real environments. With a much smaller LLM (LLaMA 3.1-8B) used in our planner, we achieve a similar success rate compared to the existing method using a much larger LLM (GPT-4o~\cite{openai2024gpt4o} or similar~\cite{openai2024gpt4o_mini}). However, in terms of efficiency, our method significantly improves upon the existing method. We save the computational requirements by 6.6 times. We successfully reduce the computation load and run our system on a local server. Largely improve the real-time performance. 

Our main contributions to this paper can be listed as follows:

\begin{itemize}
    \item To the best of our knowledge, we are the first work focusing on improving the efficiency of using LLM as the planning module for robots to finish long-horizon tasks.
    \item We propose restricting the operational scope of the robots and limiting the skill set accessible to the planner to mitigate planning complexity.
    \item We propose to close the planning and control loop, which means constantly providing visual feedback to the planner; the feedback allows the planner to change its previous planning steps.
    \item We augment the planner with a memory module, providing instant memorization of the recently used objects, further reducing the latency of repetitively referencing the same objects in a long-horizon task.
\end{itemize}

The rest of this paper is organized as follows. \Sect{sec:relate} introduces the related work. \Sect{sec:method} proposes our method. \Sect{sec:eval} demonstrates the effectiveness of our work with experiments in simulation and real scenarios. \Sect{sec:concl} concludes this paper. 
\section{Related Work}
\label{sec:relate}

\paragraph{Long-horizon Task Planning via multi-modal LLMs.} Planning ability decides the upper bound of how intelligent a robot can be. Traditional rule-based planners can enable robots to solve short and simple tasks. The emergence of large language models shows the potential for reasoning and complex task decomposition~\cite{ye2023large,ho2022large,shen2023taskbench}, and thus, it has soon been applied in robot planning. At first, researchers try to incorporate LLMs into the planning cycle of robots~\cite{kannan2023smart,ding2023task}. Soon, multi-model LLMs or VLMs take over. Among these efforts, Code-as-Policies~\cite{liang2023code}, PaLM-E~\cite{driess2023palm}, and robotic transformers~\cite{brohan2022rt,brohan2023rt,joublin2024copal} demonstrate significant improvement on the success rate on long-horizon tasks. Usually, these VLMs take language instructions such as ``clean the table'' and visual observations and generate action sequences for the robots to finish the task. 

\paragraph{End-to-end Approach and Modular-based Solutions.} Generally, the use of LLM can be classified into two categories. The first one is the end-to-end approach, which directly outputs robot actions from the models~\cite{kim2024openvla,team2024octo,black2024pi0}. End-to-end approaches must train the model from scratch or fine-tune a trained model, which requires heavy data. Modular-based solutions tend to wrap the skills robot's own into APIs and use models to program based on APIs~\cite{zhi2024closed}. Such skills include one module in the robot computing pipeline, such as navigation and perception, or a combination of multiple modules, such as exploration and grasp. 

\paragraph{Close-loop Control.} The first batch of research works on leveraging LLMs as the planning module performs open-loop control, where the robots will execute the set of instructions provided by the planner~\cite{ahn2022can,dalal2024plan}. However, replanning with feedbacks, or close-loop control is one of the key steps of enabling robots to execute long-horizon tasks~\cite{li2024closed,bu2024closed}. COME-Robot~\cite{zhi2024closed} and REPLAN~\cite{skreta2024replan} starts to leverage the LLMs they use for planning to digest visual feedbacks to help replanning. In line with prior works, we show that with appropriate feedbacks, using small models as the planner can enhance their planning capability and achieve similar success rate compared to large models. 
\section{Method}
\label{sec:method}
We describe the method we use to enhance an LLM-based planner. Specifically, the enhancement contains a limited instruction set (\Sect{sec:method:isa}), frequent visual feedback (\Sect{sec:method:fb}), and memory augmentation (\Sect{sec:method:aug}).

\subsection{Architecture}

\begin{figure}[htbp]
  \centering
  \includegraphics[width=\linewidth]{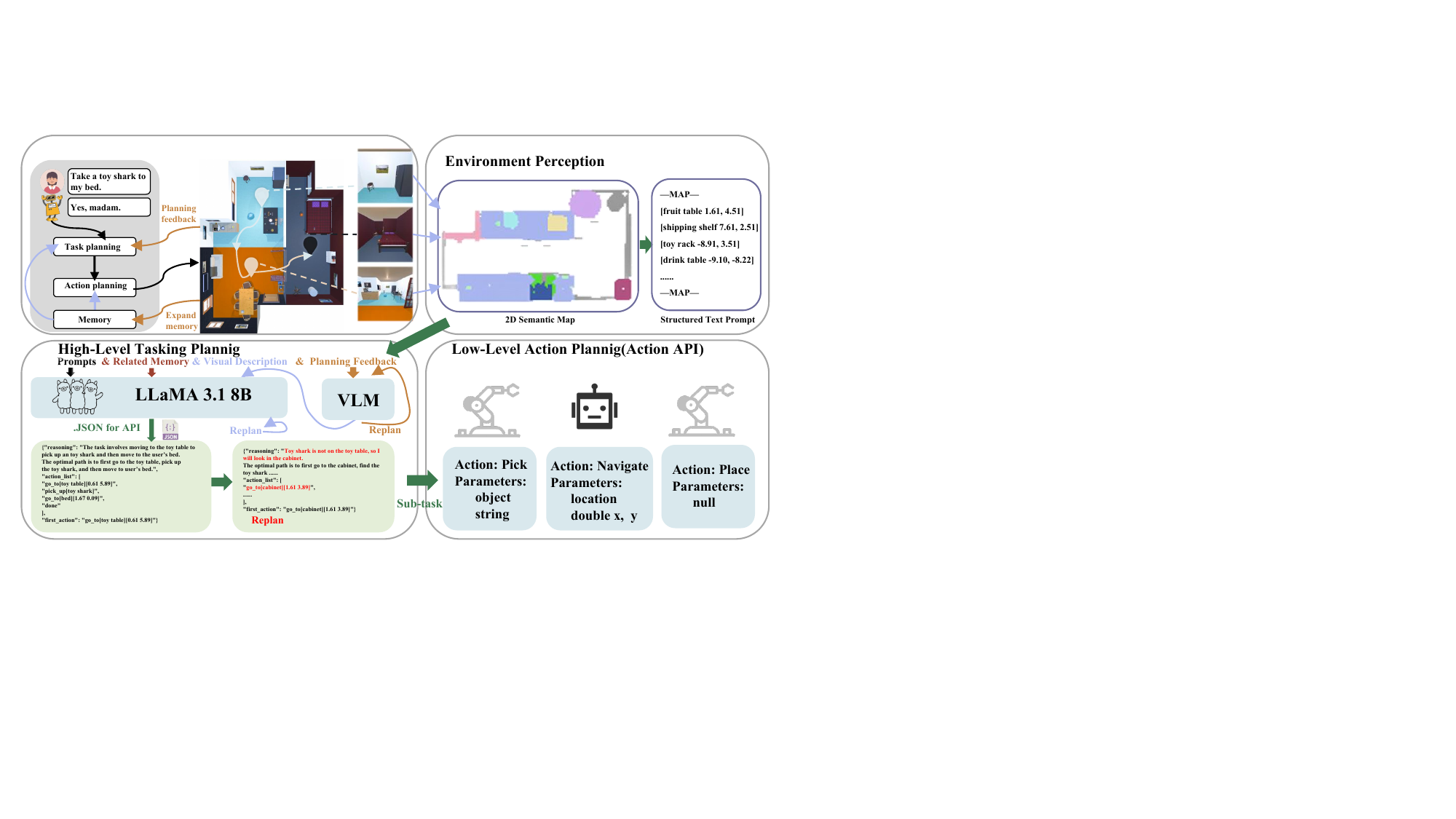}
  \caption{Architecture of DaDu-E.}
  \label{fig:main}
  \vspace{-14pt}
\end{figure}

We build three main modules in DaDu-E for robust closed-loop planning as Fig \ref{fig:main}: instruction sets, planning feedback, and memory augmentation. These modules address critical gaps in existing robotic systems, often leading to static and unreliable task execution. Instruction sets provide a structured and machine-readable representation of task instructions, overcoming the ambiguity inherent in natural language commands. This structure ensures clarity and minimizes errors during task decomposition and execution for LLM planner. Robots often fail to interpret complex or context-dependent commands without such a module, leading to task breakdowns or misinterpretations. LLM first gets instructions from the user and then breaks them down into sub-tasks for these instruction sets to action. Each instruction is blessed with its dependent running function for better performance. Planning feedback introduces a robust mechanism for real-time adaptability by enabling the system to dynamically replan actions in response to environmental changes or planning failures. This capability is essential because static planning approaches are brittle and often incapable of handling unexpected changes in dynamic environments, such as the absence of a required object or a blocked path. This feature is crucial for operating in dynamic, unpredictable environments. Finally, memory augmentation enhances the robot’s ability to reuse prior knowledge, thereby reducing latency and improving task reliability, which is especially helpful in environment-changing circumstances. Traditional systems that rely exclusively on real-time data lack the capacity to store and recall past interactions, leading to inefficient re-planning and redundancy in decision-making processes.

\subsection{Instruction Sets}
\label{sec:method:isa}
Instructions are bridges between planners and robots. Detailed and pre-programmed instructions provide encapsulation of the abilities of the robots, just as the instruction set architecture (ISA) like ARM~\cite{armv8manual} and X86~\cite{intel64_manual} provide encapsulation of the underlying hardware to the programmers. Thus, the LLM-based planner can only focus on high-level planning, guiding robots to finish long, complex tasks. 

The trend of designing low-level instruction sets is to provide as rich instructions as possible.  The design principle is to provide a huge sampling space for the LLM to pick the best chain of actions. For example, Robotic Transformer (RT-1)~\cite{brohan2022rt} provides nine skills for the planner, and COME-Robot provides six~\cite{zhi2024closed}. We find that among the six skills provided in COME-Robot, some of them overlap. For example, the skill \texttt{explore\_local} and \texttt{explore\_global} can be replaced by the skill \texttt{navigation} and a map of the environment.  

Although rich instructions can enhance the planner's search space, it also burdens the model. First, currently, the instructions are usually fed into LLMs in different forms of prompts; larger instruction sets indicate longer context, which are generally more challenging for affordable LLMs such as LLaMa with 7 billion parameters~\cite{touvron2023llama}. Second, not all the skills in larger instruction sets are frequently used. Some may even lead to longer execution time and worse planning results. Specifically, in our evaluation, we find the skill \texttt{explore} can lead to lower success path length (SPL)\cite{spl}. 

We aim to provide leaning skill sets tailored for the working environment and applications. For example, in a warehouse environment where the robots perform tasks such as organizing objects and picking objects for users, the instruction sets should be simple and contain only three skills. The \texttt{navigate} instruction guides the robot to the vicinity of the specified shelf and positions it facing the center of the shelf. As we provide a detailed semantic map for the planner, the \texttt{navigate} instruction can thus transform semantic information to geometric information. The \texttt{grasp} instruction, utilizing sensor and semantic information, directs the robot to grasp a specific object. In contrast, the instruction in the \texttt{place} controls the robot to place the object on a table in front of it. Necessary parameters and return values for each instruction are shown in detail in Table \ref{tab:functions}.

With the reduced size of instruction sets, we significantly reduce the context length. The average length of our prompts feed into the planner is only 543 tokens, significantly shorter than existing works. The planner will generate a segment of a JSON file that lists the actions to be taken to perform the long and complex task. Notice that the JSON file is not unchangeable; as we will provide frequent feedback, the action list may vary along the execution of the task. 

\begin{table}[ht]
\centering
\caption{Function Descriptions.}
\label{tab:functions}
\resizebox{0.75\columnwidth}{!}{%
\begin{tabular}{p{3cm} p{5cm} p{4cm}} 
\toprule
\textbf{Function} & \textbf{Parameters} & \textbf{Return Value} \\
\midrule
\texttt{navigate} & location \textbar\ double x, double y & result, observation \textbar\ Boolean, Image \\
\texttt{pick} & name of target object \textbar\ string & grasping result \textbar\ Boolean \\
\texttt{place} & null & place result \textbar\ Boolean \\
\bottomrule
\end{tabular}%
}
\end{table}

\subsection{Planning Feedback}
\label{sec:method:fb}
Planning feedback is another key contribution we propose. Most LLM-based planners are open-loop control, where robots only execute the instructions generated by the planner and do not provide feedback. COME-Robot is one of the first several works introducing feedback mechanism into the planner. It provides instant feedback on whether the task succeeds and related images as feedback to the system. 

We go one step further by having the planner evaluate the previous instruction and use the results as feedback for further planning. For example, once the \texttt{navigate} instruction finishes and the next instruction is \texttt{grasp}, the planner will evaluate whether the robot is at the optimal position for grasp instruction. Specifically, we feed the results of the previous instructions, the current status of the robot, the image captured at the current position, and the results of parsing the image using a Vision Language Model (VLM), such as the objects in the image and the geometric relationship. Meanwhile, we send the current state back to the LLM planner in order to get the next loop action.

With detailed feedback, the planner will be required to evaluate the previous instructions based on the feedback, consider whether to finish the next instruction and make further plans. Specifically, we ask the planner to do two things. If the planner decides the previous instruction was not properly executed, it can reasonably alter the future plan. Otherwise, the original plan is followed. For example, upon reaching the current table, if the table does not contain the designated grasping target, the LLM planner will instruct the robot to navigate to the next potential table and repeat this process until all plausible tables have been explored.

The occasions of re-planning do happen, even when the previous instruction finishes successfully. For example, when the planner issues a \texttt{navigate} instruction followed by a \texttt{grasp} instruction, the first navigation usually stays in a coarse granularity. After the robot finishes the \texttt{navigate} instruction, it usually locates itself at a position near the object to pick. However, the grasp instruction may not be successfully executed due to the geometric distance between the robot and the object (i.e., a large table with an irregular shape). Under such circumstances, the planner should and will change the rest of the plan, specifically by changing the next instruction from \texttt{grasp} into \texttt{navigate} to a position closer to the object. 

The robot must navigate to the optimal table side for grasping. For instance, if the target "apple" is on the front side and the robot is on the back, the distance exceeds the arm's working range. To address this, we capture the current table image upon reaching the navigation goal and send it to the VLM planner based on LLaVA-OneVision-8B\cite{li2024llava}, which identifies the correct table side as the final navigation goal.

\subsection{Memory Augmentation}
\label{sec:method:aug}

Our robot system's memory module denoted $M$, consists of two main components: short-term memory $M_S$ and long-term memory $M_L$. Long-term memory maintains a semantic map of the environment, functioning similarly to semantic memory in the human brain. Short-term memory, by contrast, focuses on frequently changing information, such as the states and the positions of objects, resembling episodic memory in the brain. 

\paragraph{Short-term memory}
The short-term memory $M_S$ is represented as follows:
\begin{equation}
M_S = \{ M_{S_{1}}, M_{S_{2}}, ..., M_{S_{N}} \}
\end{equation}
where each $M_{S_i}(1 \leq i \leq N)$ is a short-term memory unit, and $N$ denotes the total number of units. 

The short-term memory module is organized as structured textual information, including object categories, locations, and image summaries. Each memory unit is assigned a unique ID to track object recurrence. When an object reappears, a new unit replaces the previous one. A single short-term memory unit is represented as: 
\begin{equation}
M_{S_i} = \{ID,OBJECT,POSITION,IMG\}
\end{equation}
where $OBJECT$ refers to the object category, $POSITION$ denotes the object's location, and $IMG$ represents a summary generated by the vision-language model (VLM) based on the object's image input.

When the robot needs to recall short-term memory, it uses semantic similarity to find the most relevant memory unit. Each $M_{S_i}$ is embedded as a vector set, and cosine similarity is calculated between the current Instructions $I$’s vector and stored memory vectors, retrieving the most similar unit $M_{S^*}$:
\begin{equation}
M_{S_{i}}^{*} = \arg\max_i \left( \text{Sim}_{COS}\left( f(I), \left\{ f(M_{S_{i}}) \right\}_{i=1}^N \right) \right)
\end{equation}
where $M_{S_{i}}^{*}$ is the memory unit most similar to the current instruction $I$, $f(\cdot)$ is a normal text embedding model, $\text{Sim}_{COS}(A,B)$ denotes the cosine similarity between A and B.

Based on the retrieved memory unit, the planner can dynamically adjust the parameters of action functions in the JSON file. For example, if at time \( t_0 \), the robot completes the task “place all yellow fruits on the storage rack,” the planner may not accurately retain all prior task details. Later, at time \( t_3 \), when given the instruction “find a yellow fruit and place it on the dining table,” the planner might generate an incorrect action list based on outdated information, mistakenly assuming the yellow fruits are still on the table.

\paragraph{Long-term memory}
Long-term memory stores persistent information about the scene, such as obstacle distribution, fixed object locations, and scene layout. During task execution, it primarily assists with global path planning and provides a deeper scene understanding when required.
This module is structured in two layers: a grid map at the lower level, which represents obstacles in the scene and is used for basic navigation and path planning, and an upper layer with semantic text labels corresponding to grid coordinates. These labels are used as prompts for the large language model (LLM), aiding the robot in understanding its environment.

\section{Evaluation}
\label{sec:eval}

\subsection{Evaluation Environment set-up}

\lstdefinelanguage{JSON}{
    basicstyle=\ttfamily\small,
    numbers=left,
    numberstyle=\tiny\color{gray},
    stepnumber=1,
    numbersep=5pt,
    showstringspaces=false,
    breaklines=true,
    frame=lines,
    backgroundcolor=\color{white},
    keywordstyle=\color{blue},
    stringstyle=\color{red},
    morestring=[b]",
    literate=
     *{0}{{{\color{orange}0}}}{1}
      {1}{{{\color{orange}1}}}{1}
      {2}{{{\color{orange}2}}}{1}
      {3}{{{\color{orange}3}}}{1}
      {4}{{{\color{orange}4}}}{1}
      {5}{{{\color{orange}5}}}{1}
      {6}{{{\color{orange}6}}}{1}
      {7}{{{\color{orange}7}}}{1}
      {8}{{{\color{orange}8}}}{1}
      {9}{{{\color{orange}9}}}{1}
      {:}{{{\color{blue}:}}}{1}
      {,}{{{\color{blue},}}}{1}
      {\{}{{{\color{black}\{}}}{1}
      {\}}{{{\color{black}\}}}}{1}
      {[}{{{\color{black}[}}}{1}
      {]}{{{\color{black}]}}}{1},
}

We conducted an evaluation of DaDu-E in both a real-world environment and a 1:1 simulation environment.

The real-world experimental environment was carefully designed to simulate a variety of scenarios that align with the defined task levels, as summarized in Table~\ref{tab:real_world}. The setup consists of multiple marked locations, each serving distinct functions in embodied intelligence testing. These include a Storage Rack, a Fruit Table, a Drinks Table, a Toys Table, and a User Entry Point. Each location was chosen to evaluate specific robotic skills such as object recognition, grasping, classification, dexterity, and user interaction.

\begin{figure}[htbp]
  \centering
  \includegraphics[width=0.6\linewidth]{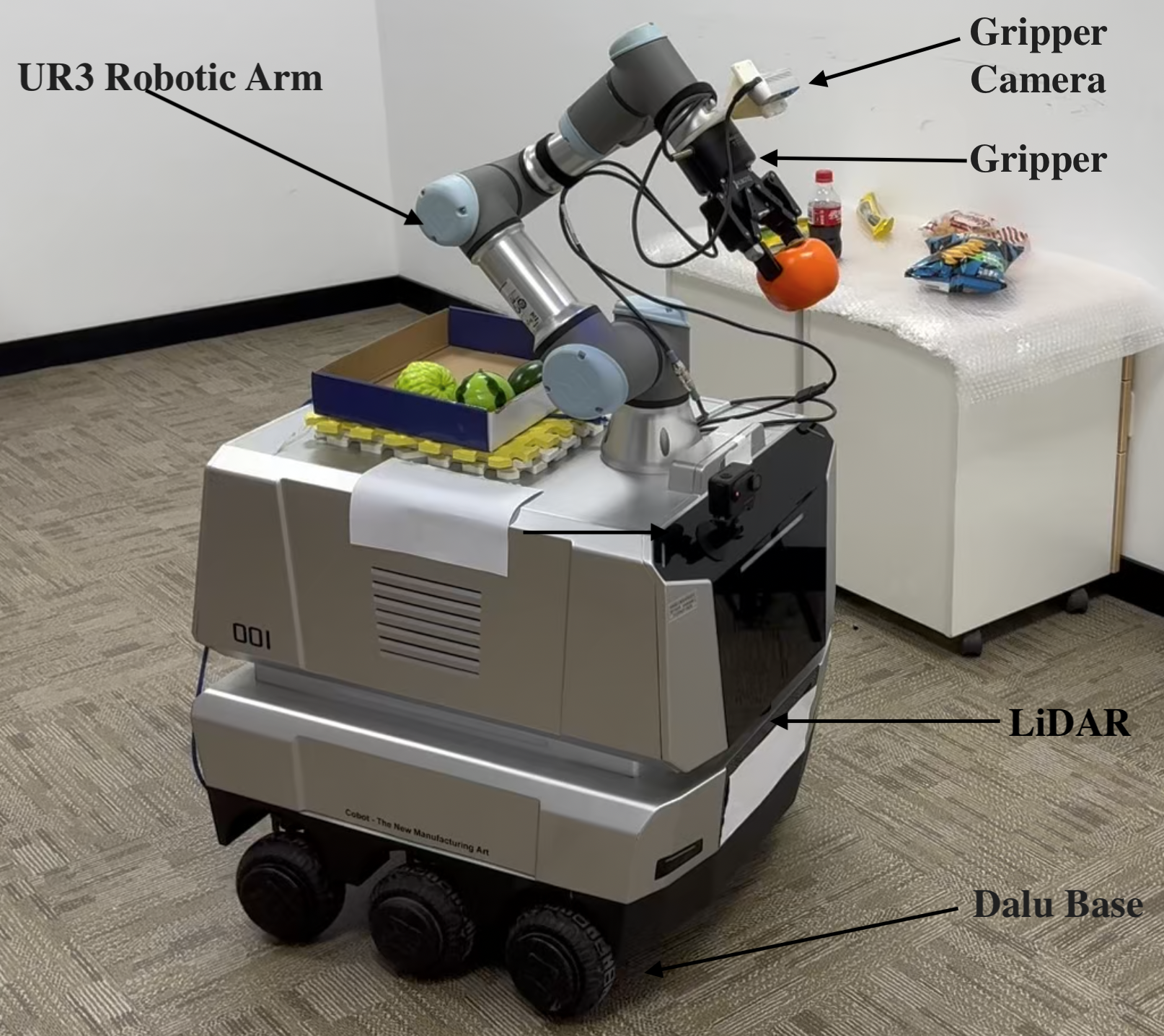}
  \caption{Overview of our real robot.}
  \label{fig:robot}
\end{figure}

For the real-world environment, the robotic hardware utilized consists of a Dalu mobile base \cite{dalurobot_stepper_robot_base} and a UR3 robotic arm \cite{UR03}. Both setups were equipped with basic hardware components, including two-finger Robotiq grippers\cite{robotiq_adaptive_grippers}, Intel d435 depth cameras\cite{intel_d435}, and RPLIDAR A2 LiDAR sensors\cite{lidar}, to ensure the robots could perform mobile manipulation tasks effectively. We build the sim and real environment on Ubuntu 20.04 and ROS noetic. We set the depth camera on the gripper for a better grasping view. The construction of our robot is in Fig \ref{fig:robot}, and robot hardware information details can reference Table \ref{tab:robot hw}. Our experiments were conducted using a mobile manipulator robot. All the model inference tasks are based on a single NVIDIA 3090 GPU with 24G GPU memory\cite{3090}.

The diverse setup provided a comprehensive testing ground for the robot's capabilities across all task levels, ensuring scalability and adaptability in real-world scenarios.\ref{tab:real_world}.

\begin{table}[h]
\centering
\caption{Robot hardware configuration.}
\label{tab:robot hw}
\begin{tabular}{cc}
\toprule
\textbf{Hardware} & \textbf{Setting} \\ 
\midrule
DoFs of Robot Arm & 6 \\ 
DoFs of Gripper & 1 \\ 
Wheels of Base & 6 \\ 
Depth Camera & 1 \\ 
LiDAR & 1 \\ 
Camera Resolution & 640*480 \\ 
\bottomrule
\end{tabular}
\end{table}

\begin{table*}[h!]
\centering
\label{fig: robot}
\caption{Functions of Marked Locations in Embodied AI Environment Testing.}
\label{tab:real_world}
\begin{tabular}{p{3cm} p{9cm}}
\toprule
\textbf{Location} & \textbf{Function in Embodied Intelligence Testing} \\ 
\midrule
Storage Rack & Placing items that need to be sorted or classified, simulating the robot's ability to organize and classify items in different tasks. \\ 
Fruit Table & Placing fruits, testing the robot's ability to recognize, grasp, and transport light objects like fruits. \\ 
Drinks Table & Placing drinks, testing the robot's ability to recognize, grasp, and transport heavier objects like drink bottles. \\ 
Toys Table & Placing toys, testing the robot's dexterity and accuracy in handling small or irregularly shaped objects. \\ 
User Entry Point & Simulating the starting or ending point of user-robot interaction, such as initiating tasks or reporting after task completion. \\ 
\bottomrule
\end{tabular}
\end{table*}

For the simulation environment, we utilize a Gazebo simulator based on our real-world environment because of its exceptional generalizability. This allows for seamless communication between simulated components and real-world robots. 


\subsection{Baseline methods}
\label{sec: baseline}
To evaluate task-finishing ability, we adopt Code as Policies (CaP) \cite{liang2023code}, a state-of-the-art approach that demonstrates strong performance in task execution. The CaP method utilizes LLM to generate code that directs robots to execute complex tasks. This approach shares similarities with our method, which involves the design of code APIs, task decomposition, and the establishment of a memory mechanism. Thus, the CaP method can serve as a compelling baseline to evaluate the reliability of our approach. Meanwhile, as the first state-of-the-art robotic closed-loop computing pipeline, COME-robot\cite{zhi2024closed} provides a long-horizon and robust task-finishing ability. However, due to the unavailability of the open-source implementation of COME-Robot, we reconstruct COME-Robot locally with the same planning method with our designed APIs to build a strong pipeline for evaluation, referred to as \textbf{COME-Robot*}. Similarly, for CaP works as on table-on tasks instead of mobile manipulate works, we take CaP as \textbf{CaP*} with the same eval method in \cite{zhi2024closed} and take GPT-3.5 as its planner.

To assess computational cost, we utilize RT-2 \cite{brohan2023rt} ,PaLM-E\cite{driess2023palm} and RoboFlamingo \cite{li2023vision} as benchmark models. These state-of-the-art end-to-end embodied AI frameworks are notable for their advanced capabilities in high-level task comprehension and planning. By conducting a comparative analysis against these baselines, we aim to demonstrate the computational efficiency and performance trade-offs of our proposed approach within the domain of embodied intelligence.

\subsection{Embodied AI task categorization}
\label{sec: task design}



We generally design four levels of tasks to systematically evaluate the robot’s capabilities at increasing levels of complexity, aiming to demonstrate the progression from basic functionalities to advanced decision-making. This categorization provides a structured framework for analysis, enabling clear insights into the robot’s strengths and areas for improvement while highlighting the benefits of a gradual increase in task difficulty for performance benchmarking. The appendix provides detailed task design.

\lstset{
  basicstyle=\ttfamily\footnotesize, 
  frame=single,                      
  numbers=left,                      
  numberstyle=\tiny,                 
  breaklines=true, 
  keywordstyle=\color{blue}\bfseries, 
  stringstyle=\color{red},
  commentstyle=\color{green!50!black}, 
  showstringspaces=false
}

\paragraph{Level 1: Basic Task Execution Level}
At this level, robots are tasked with performing fundamental sub-tasks such as grasping and placing various objects. The primary objective of this level is to validate the robot's success rate in different object manipulation and navigation tasks. Tasks typically involve specific object operations, such as grasping a strawberry and placing it on a toy table or picking up a squirrel toy and delivering it to a shipping table. These tasks emphasize the robot's basic operational capabilities and elementary environmental adaptability, laying the groundwork for more advanced tasks. We design 10 different tasks of level 1; four of the example instructions from users are as Table \ref{tab:instructions 1}.

\begin{table}[h!]
\centering
\caption{Example Instructions For Task Level 1.}
\label{tab:instructions 1}
\begin{tabular}{@{}c@{}}
\toprule
\textbf{Instruction} \\ 
\midrule
Grasp a strawberry and put it on toy table. \\
Find delicious fenta can and place on shipping table. \\
Move to the fruit table. \\
Pick up a strawberry and put it on shipping table. \\
\bottomrule
\end{tabular}
\end{table}

\paragraph{Level 2: Multi Tasks Execution Level}
The Advanced Task Execution level involves multi-step and composite operations, requiring robots to possess enhanced task processing and environmental adaptation capabilities. Tasks at this level typically encompass multiple operations, such as identifying target objects, grasping, moving, and placing them in specified locations. The complexity and diversity of task steps are key characteristics of this level. For example, a robot may need to avoid obstacles or adjust its operational strategy while performing a grasping task in response to environmental changes. This level evaluates the robot's performance in more complex and dynamic environments. We design 7 different tasks for level 2, four of the example instructions from users are as follows:

\begin{tcolorbox}[title=Task 1]
Grasp a Pepsi can and place it on the fruit table. Pick up a squirrel toy and place it on the shipping table. Pick a Sprite can and put it on the fruit table.
\end{tcolorbox}
\begin{tcolorbox}[title=Task 2]
Find a plum and place it on the shipping table. Move a strawberry to the toy table. Set the squirrel toy on the fruit table.
\end{tcolorbox}
\begin{tcolorbox}[title=Task 3]
Distinguish a squirrel toy and put it on the shipping table. Find another squirrel toy and put it on the fruit table. Grasp a toy shark and put it on the drink table.
\end{tcolorbox}
\begin{tcolorbox}[title=Task 4]
Locate the shark toy and position it on the fruit table. Fetch the tea box and place it on the same fruit table. Discover the ladybug toy and set it on the drink table.
\end{tcolorbox}

\paragraph{Level 3: Autonomous Decision-Making Level}
The Autonomous Decision-Making level demands that robots have the capability to make independent decisions in dynamic environments. Tasks at this level extend beyond executing predefined operational steps, requiring robots to adjust based on real-time feedback and environmental changes. For instance, robots may need to reposition target objects, optimize their paths, or avoid new obstacles. This level emphasizes the robot's ability to perform situational awareness and task optimization in uncertain environments, showcasing the robot's intelligent decision-making capabilities. We design 10 different tasks of level 3, four of the example instructions from users are as Table \ref{tab:instructions 3}.

\begin{table}[h!]
\caption{Example Instructions For Task Level 3.}
\label{tab:instructions 3}
\centering
\begin{tabular}{@{}c@{}}
\toprule
\textbf{Instruction} \\ 
\midrule
Pick up a yellow fruit to shipping table. \\
Put the blue can drink to shipping table. \\
Help me get the biggest toy and place it on shipping table. \\
Sorting the apple, coke from purchase table and send it to the reasonable place. \\
\bottomrule
\end{tabular}
\end{table}

\paragraph{Level 4: High-Level Cognition Level}
The High-Level Cognition level represents the pinnacle of robotic technology development, requiring robots to possess advanced cognitive and decision-making abilities similar to humans. This level involves natural language understanding, multi-modal information integration, and high-level reasoning. Tasks typically require robots to comprehend complex natural language instructions and execute multi-step tasks in complex environments while handling unexpected situations and correcting errors. For example, a robot might need to complete a series of object grasping and placing tasks based on natural language high-level unclear reference commands and adapt its strategy during task execution, such as "Completed the receiving of all the objects" or "I need a pick-me-up drink, complete its shipment". This level evaluates the robot's advanced cognitive abilities and comprehensive intelligence. We design 10 different tasks of level 4, four of the example instructions from users are as Table \ref{tab:instructions 4}.

\begin{table}[h!]
\centering
\caption{Example Instructions For Task Level 4.}
\label{tab:instructions 4}
\begin{tabular}{@{}c@{}}
\toprule
\textbf{Instruction} \\ 
\midrule
I am hungry, ship me something to eat. \\
Give me a fruit rich in vitamin c, complete its shipment. \\
Completed the receiving of apples. \\
Completed the receiving of all the drink. \\
\bottomrule
\end{tabular}
\end{table}

\subsection{Evaluation analysis}

\subsubsection{Computing cost}
Table \ref{tab:model_pipeline} presents the model parameters and FLOPs (Floating Point Operations) associated with different components of DaDu-E. The pipeline consists of the Llama3.1-8B\cite{dubey2024llama} as planner LLM, ROS move-base package\cite{move_base} as Navigation module, AnyGrasp\cite{fang2023anygrasp}, LangSAM\cite{lsa} and UR RTDE\cite{ur_rtde} as Grasp module, and llava-onevision-qwen2-7b-ov\cite{li2024llava} as Vision-Language Model (VLM). The place module is a rule-based method based on RTDE. As a result, its computing cost is much less.  The number of input tokens varies across different scenarios, depending on the length of input words at the classes of prompting, first feedback(fb1), and second feedback(fb2) stages. This variation arises from our prompt configuration, as detailed in \Sect{sec:append}.

\begin{table}[ht]
\centering
\caption{DaDu-E's Model Parameters and FLOPs.}
\label{tab:model_pipeline}
\begin{tabular}{@{}ccccccccc@{}}
\toprule
\textbf{} & \multicolumn{3}{c}{\textbf{LLM}} & \textbf{Navi} & \multicolumn{2}{c}{\textbf{VLM}} & \textbf{Grasp} & \textbf{Place} \\ 
\cmidrule(r){2-4} \cmidrule(lr){6-7} 
\textbf{} & \textbf{prompt} & \textbf{fb1} & \textbf{fb2} & \textbf{} & \textbf{fb1} & \textbf{fb2} & \textbf{} & \textbf{} \\ 
\midrule
\textbf{token num} & 543 & 22 & 3 & / & 59+196 & 15+196 & / & / \\ 
\textbf{parm num} & 8000M & 8000M & 8000M & 18.03M & 7000M & 7000M & 564M & 21 \\ 
\textbf{FLOPs}     & 26064G & 1056G & 144G & 0.002G & 10710G & 8862G & 837.6G & 27 \\ 
\bottomrule
\end{tabular}
\end{table}

The computational cost of this configuration is detailed in Table~\ref{tab:model_params}. DaDu-E demonstrates remarkable computational efficiency compared to state-of-the-art models such as RT-2\cite{brohan2023rt}, PaLM-E\cite{driess2023palm}, and RoboFlamingo\cite{li2023vision}. We estimate the total computing cost of DaDu-E as \ref{eq:cost}.
\begin{equation}
\label{eq:cost}
\text{cost} = c_{\text{plan}} + c_{\text{navi}} + c_{\text{vlm}} + c_{\text{grasp}} + c_{\text{place}}
\end{equation}

\begin{figure}[htbp]
  \centering
  \includegraphics[width=1\linewidth]{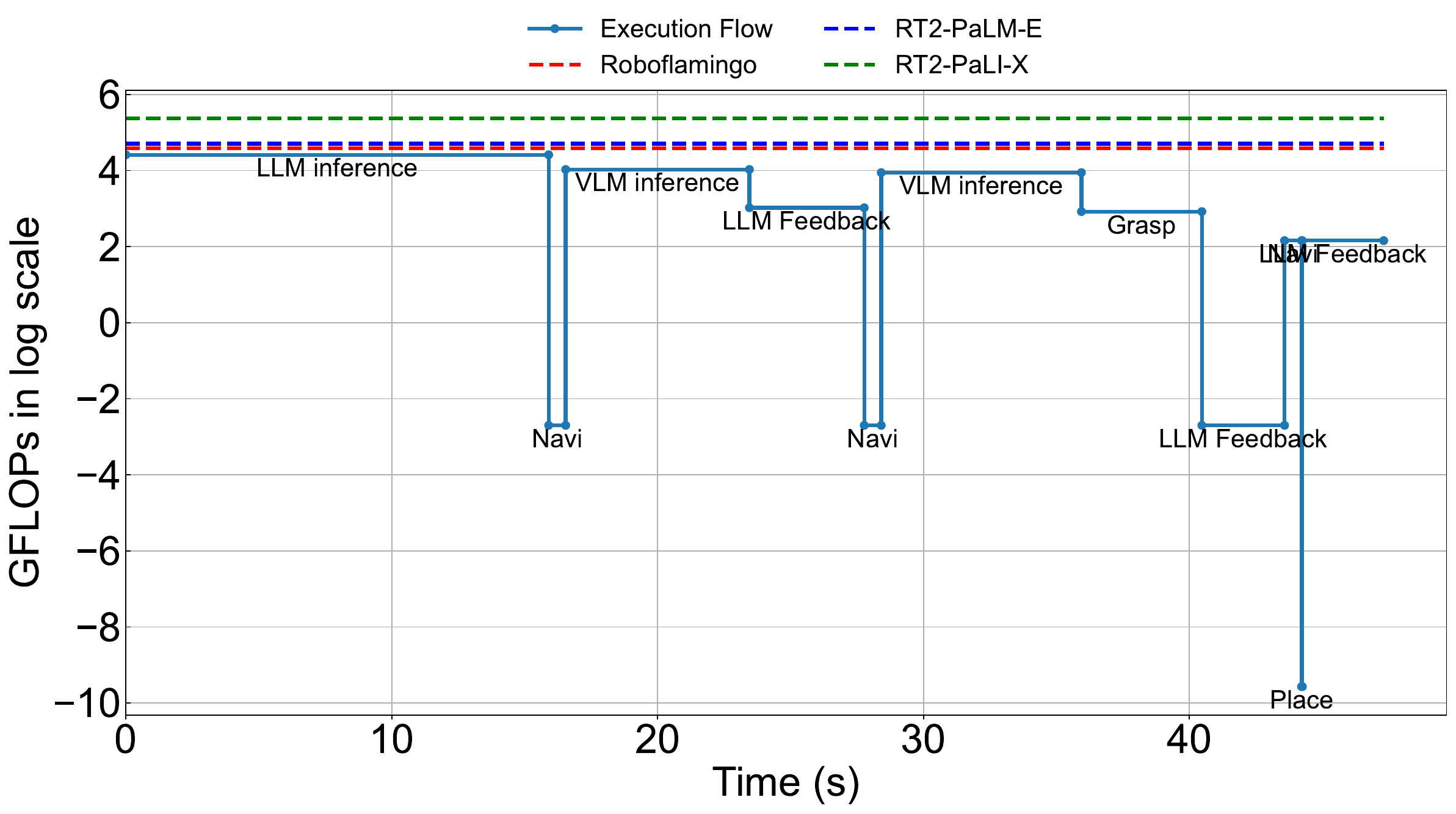}
  \caption{Overview of computing cost with baselines.}
  \label{fig:flops_change}
\end{figure}

With the peak of only \textbf{8 billion parameters}, DaDu-E significantly reduces the model size while maintaining robust task performance. In contrast, RT-2\cite{brohan2023rt} (based on PaLM-E and PaLM-X) utilizes 12 billion and 55 billion parameters, respectively, and PaLM-E reaches an immense 562 billion parameters, leading to substantially higher computational demands. Even RoboFlamingo, which operates in configurations ranging from 3 billion to 9 billion parameters, cannot consistently match the computational efficiency of our approach. 

\begin{figure}[htbp]
  \centering
  \includegraphics[width=0.6\linewidth]{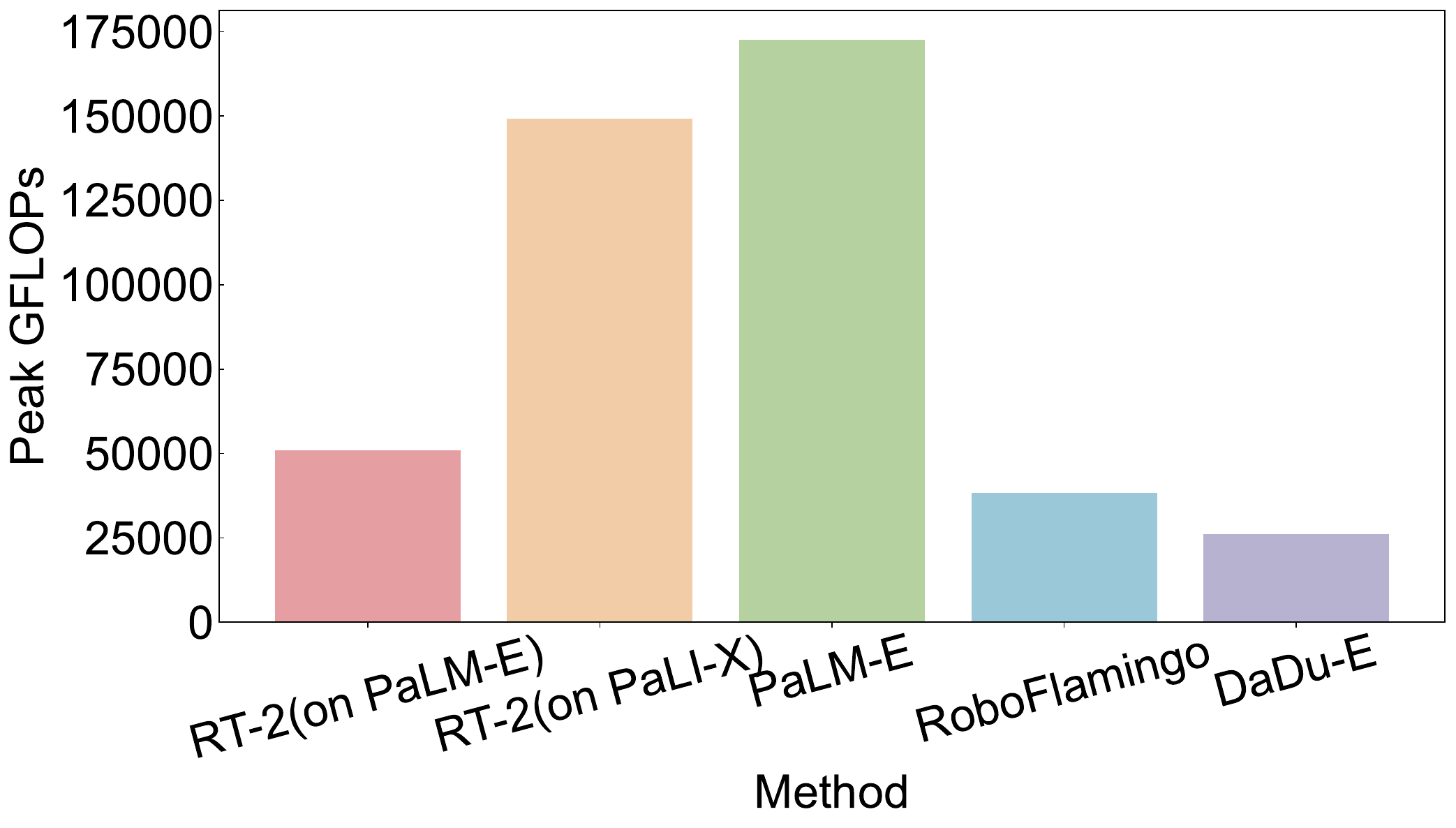}
  \caption{Overview of peak cost with end-to-end methods.}
  \label{fig:flops}
\end{figure}

Fig \ref{fig:flops_change} shows the change of computing cost during a single task example, "Take a banana to shipping table." comparing with end-to-end methods and CaP*. The maximum computational requirement of DaDu-E during LLaMA inference is \textbf{26064 GFLOPs}, which has been the lowest among all compared models in Tab \ref{tab:model_params}. The computational cost during LLM feedback is less than LLM inference due to the number of input tokens at the feedback stage. Meanwhile, RT-2 requires 50976 GFLOPs to 49160 GFLOPs depending on its configuration, with PaLM-E imposing an extraordinary computational burden of 172646.4 GFLOPs. The RoboFlamingo\cite{li2023vision} method depending on its parameter size, incurs 281000 GFLOPs, exceeding our pipeline in larger configurations. This illustrates that DaDu-E not only reduces model size but also minimizes computation, enabling resource-efficient performance. Additionally, DaDu-E uses an average input token size of \textbf{543} for a prompt and \textbf{196} for a picture, avoiding using too many examples in the input prompt with 1024+196 as CaP*. This further reduces computational overhead while ensuring the effectiveness of task comprehension and execution.

Fig \ref{fig:flops} presents a comparison of the peak computational costs required for completing a single task of “Take a banana to the shipping table” between our method and several baselines. Notably, the maximum FLOPs required by our pipeline during LLaMA inference is \textbf{26064 GFLOPs}, counting the lowest among all compared methods. In contrast, RT-2 incurs a computational cost ranging from 49160 GFLOPs to 50976 GFLOPs, depending on its dataset configuration, while PaLM-E imposes a significantly higher burden of 172646.4 GFLOPs. The RoboFlamingo\cite{li2023vision} approach requires up to 281000 GFLOPs, underscoring the efficiency of our pipeline in achieving competitive performance with significantly lower computational demands. Meanwhile, our peak cost is much less than CaP* with GPT-3.5 and a larger token number in Table \ref{tab:model_params}.

\begin{table*}[htbp]
    \centering
    \caption{Model params and costs.}
    \label{tab:model_params}
    \begin{tabular}{lccc}
        \toprule
        \textbf{model} & \textbf{params} & \textbf{avg input token number} & \textbf{peak FLOPs} \\
        \midrule
        RoboFlamingo & 9B & 32+196 &  38232G \\
        RT-2 (on PaLM-E) & 12B & 32+196 & 50976G \\
        RT-2 (on PaLI-X) & 55B & 32+196 & 149160G \\
        PaLM-E & 562B & 32 & 172646.4G \\
        CaP* & 175B & 1024+196 & 1281000G \\
        \rowcolor{mygray}DaDu-E & 8.60B(total) & 543+196 & 26064G\\ 
        \bottomrule
    \end{tabular}
\end{table*}

\subsubsection{System performance}
In this section, we statistics the success rate(SR) of different levels of planning execution. Table \ref{tab: performance} shows the performance of CaP*, COME-Robot*, and DaDu-E across 20 tasks at each level. Based on the performance results across different task levels, we evaluate our approach against two baseline methods: COME-Robot* and CaP*, across four task complexity levels. The “Ideal” column represents the theoretically optimal performance, while the “Execute” column reflects actual observed performance.

At Levels 1 and 2, our method achieves success rates(SR) of 0.90 and 0.70, respectively. These results match the performance of COME-Robot* and surpass CaP* by 2.1 $\times$. In Level 3 tasks, our method demonstrates a success rate 2.6 $\times$ higher than CaP* but falls short of COME-Robot* by 1.6 $\times$. At Level 4, our method matches COME-Robot*’s performance and significantly outperformed CaP*. These findings demonstrate that our method reliably handles tasks requiring foundational skills, such as object manipulation and simple navigation, achieving state-of-the-art (SOTA) performance. However, in more complex and dynamic environments, our method exhibits slight disadvantages compared to COME-Robot* but still far surpasses CaP*.
A key advantage of our approach is its efficiency: our model uses only 0.008 times the parameters of COME-Robot* while achieving nearly comparable performance. This underscores its capability to deliver competitive results with minimal computational resources.
\begin{table}[h]
\centering
\caption{Performance Across Different Task Levels.}
\label{tab: performance}
\resizebox{1\columnwidth}{!}{%
\begin{tabular}{lcccccccc}
\toprule
\textbf{Task Level} & \multicolumn{2}{c}{\textbf{1}} & \multicolumn{2}{c}{\textbf{2}} & \multicolumn{2}{c}{\textbf{3}} & \multicolumn{2}{c}{\textbf{4}} \\
\cmidrule(r){2-3} \cmidrule(r){4-5} \cmidrule(r){6-7} \cmidrule(r){8-9}
 & \textbf{Ideal} & \textbf{Execute} & \textbf{Ideal} & \textbf{Execute} & \textbf{Ideal} & \textbf{Execute} & \textbf{Ideal} & \textbf{Execute} \\
\midrule
COME-Robot* & 1 & 0.90 & 1 & 0.70 & \textbf{1} & \textbf{0.85} & 1 & 0.95 \\
CaP* & 0.70 & 0.43 & 0.50 & 0 & 0.40 & 0.20 & 0 & 0 \\
\rowcolor{mygray}DaDu-E & \textbf{1} & \textbf{0.90} & \textbf{1} & \textbf{0.70} & 0.60 & 0.51 & \textbf{1} & \textbf{0.95} \\
\bottomrule
\end{tabular}%
}
\end{table}

Meanwhile, for testing navigation performance, we present the Success-weighted Path Length (SPL) index as a key metric to evaluate the navigation performance of our pipeline, following the methodology in \cite{spl}. Different from \cite{spl}, our SPL metric combines factors of path efficiency and task broken-down performance, offering a robust evaluation of the system’s planning and navigation performance in task execution. Specifically, SPL is calculated using the navigation distance computed in ROS, the ideal navigation distance obtained from the Dijkstra algorithm in ROS. This approach enables an assessment that accounts for both the optimality of the path taken and the likelihood of successful task completion, reflecting the system’s overall efficiency and reliability in complex task environments.

\begin{equation}
\text{spl} = \frac{1}{N} \sum_{i=1}^N S_i \frac{\ell_i}{\max(p_i, \ell_i)}
\end{equation}
Where N represents the total number of tasks from LLM planer, $S_i$ denotes the success rate for the i-th task, $l_i$ represents the shortest path length, $p_i$ indicates the actual path length taken by the robot to complete the task.

Table \ref{tab: spl} presents a comparison of the SPL metric between our method and the baseline, COME-Robot*. Across all task levels, our method matches the SPL performance of COME-Robot*. Notably, this efficiency is achieved with a model that is 125 $\times$ smaller than COME-Robot*, highlighting the effectiveness of our approach in delivering high navigation efficiency with significantly reduced complexity.

The comparable SPL values between our approach and COME-Robot across all task levels indicate similar performance in path optimality and task success. This parity in performance suggests that our pipeline maintains high navigation effectiveness without compromising task success rates, effectively balancing path efficiency and task completion under various conditions. Thus, our approach proves to be as reliable and effective as the baseline, COME-Robot*, for navigation tasks requiring both efficiency and high success rates.

\begin{table}[h!]
\centering
\caption{SPL Comparison with COME-Robot.}
\label{tab: spl}
\begin{tabular}{ccccc}
\toprule
\textbf{Task Level} & \textbf{Condition} & \textbf{N} & \textbf{SPL} & \textbf{Diff.} \\
\midrule
\multirow{2}{*}{1} & COME-Robot* & 2 & 0.9406 & 0 \\
                   & DaDu-E    & 2 & 0.9406 & - \\
\midrule
\multirow{2}{*}{2} & COME-Robot* & 5 & 0.8764 & 0 \\
                   & DaDu-E    & 5 & 0.8764 & - \\
\midrule
\multirow{2}{*}{3} & COME-Robot* & 6 & 0.9148 & 0 \\
                   & DaDu-E    & 6 & 0.9148 & - \\
\midrule
\multirow{2}{*}{4} & COME-Robot* & 4 & 0.8627 & 0 \\
                   & DaDu-E    & 4 & 0.8627 & - \\
\bottomrule
\end{tabular}
\end{table}

In summary, our method performs competitively against COME-Robot*, particularly in high-complexity tasks. It significantly outperforms CaP* across all levels, making it a viable solution for task execution scenarios requiring both reliability and adaptability.

As shown in Table \ref{tab:llmtest}, in most cases, although our planner has a significantly smaller parameter scale, it achieves the same task execution performance and as the GPT-4o planner used in COME-Robot*, and much better than CaP*. Therefore, we believe that our system greatly reduces the computational cost of embodied intelligence workflows while maintaining the same level of execution performance.

\begin{table}[h!]
\centering
\caption{Comparison of Pipeline Planners.}
\label{tab:llmtest}
\begin{tabular}{lcc}
\toprule
\textbf{System} & \textbf{Planner} & \textbf{Parameters} \\
\midrule
COME-Robot* & GPT-4o & Over 1000B \\
CaP* & GPT-3.5 & Over 175B\cite{brown2020language} \\
\rowcolor{mygray}DaDu-E & LLaMA 3.1 & \textbf{8B} \\
\bottomrule
\end{tabular}
\end{table}

\subsubsection{Latency}
\label{sec:latency}
To evaluate the latency of task execution, we analyze the time required for each task by dividing it into distinct components: planning ($t_{planner}$), navigation ($t_{navi}$), Vision-Language Model processing ($t_{vlm}$), grasping ($t_{grasp}$), and placing ($t_{place}$). Table \ref{tab:latency_comp} provides a detailed breakdown of these latency components at different task levels, facilitating a comprehensive comparison between the baseline method, COME-Robot, and our proposed approach. This evaluation aims to demonstrate the efficiency of our method across critical stages of task execution.

\begin{equation}
\text{latency} = t_{\text{planner}} + t_{\text{navi}} + t_{\text{vlm}} + t_{\text{grasp}} + t_{\text{place}}
\end{equation}

Table \ref{tab:latency_comp} compares the latency of various modules in our method with the baseline across different task levels. Our method consistently demonstrates lower total latency than the baseline at all task levels. Furthermore, as task complexity increases, the latency reduction achieved by our approach becomes more significant, improving by 2.4 $\times$ to 5.2 $\times$ compared to COME-Robot*.

A detailed analysis reveals that at task levels 1 and 4, nearly all the latency reduction in our method is attributable to the VLM component. At task levels 2 and 3, the VLM component contributes 69 $\%$ of the overall latency reduction relative to the baseline.

These findings underscore the effectiveness of our VLM processing and optimized task planning in reducing system latency. By integrating llava-onevision-7b into the data pipeline, we achieve substantial computational efficiency while maintaining performance on par with GPT-4o.

\begin{table*}[htbp]
    \centering
    \caption{Latency Comparison of Different Tasks.}
    \label{tab:latency_comp}
    \begin{tabular}{@{}cccccccc@{}}
    \toprule
    \textbf{Task Level} & \textbf{Method} & \textbf{Plan} & \textbf{Navi} & \textbf{VLM} & \textbf{Grasp} & \textbf{Place} & \textbf{Total} \\ 
    \midrule
    \multirow{2}{*}{1} 
    & COME-Robot & 15.08 & 212.04 & 26.28 & 230.11 & 43.91 & 527.42 \\
    & DaDu-E & 15.90 & 212.04 & \textbf{20.82} & 230.17 & 43.92 & \textbf{522.85} \\
    \midrule
    \multirow{2}{*}{2} 
    & COME-Robot & 33.97 & 578.99 & 78.84 & 519.57 & 131.73 & 1343.10 \\
    & DaDu-E & 26.78 & 578.99 & \textbf{62.46} & 519.57 & 131.73 & \textbf{1319.53} \\
    \midrule
    \multirow{2}{*}{3} 
    & COME-Robot & 33.97 & 737.72 & 78.84 & 519.57 & 131.73 & 1501.39 \\
    & DaDu-E & 26.78 & 737.72 & \textbf{62.46} & 519.57 & 131.73 & \textbf{1478.26} \\
    \midrule
    \multirow{2}{*}{4} 
    & COME-Robot & 22.96 & 558.34 & 52.56 & 346.38 & 88.34 & 1067.58 \\
    & DaDu-E & 22.88 & 558.34 & \textbf{41.64} & 346.38 & 88.34 & \textbf{1057.58} \\
    \bottomrule
    \end{tabular}
\end{table*}

\subsection{Ablation study}
\label{sec: ablation}
\subsubsection{Feedback}
To evaluate the impact of feedback on task execution performance, we conducted an ablation study, comparing task success rates with and without the feedback module across four task levels. In the feedback condition, the robot receives real-time feedback throughout task execution, enabling it to adjust actions based on dynamic environmental changes and task requirements. In contrast, the no-feedback condition removes this capability, assessing the robot’s performance in the execution period. This experimental design allows us to isolate and analyze the contribution of feedback to the overall task success, particularly in complex tasks requiring environmental adaptability.

Table \ref{tab: task_execution_feedback} summarizes the results of an ablation study on the feedback module. Incorporating the feedback module leads to significant improvements in task success rates compared to the method without it, with increases of 1.8$\times$, 12.7$\times$, 3.4$\times$, and 2.1$\times$ at task levels 1, 2, 3, and 4, respectively.

These findings highlight the critical role of the feedback module in enhancing the robot’s robustness. By addressing unexpected obstacles and adapting to complex, dynamic environments, the feedback mechanism significantly boosts task completion rates. The most substantial improvement was observed at task level 2, suggesting that during the planning of longer sequences involving multiple objects, LLMs require timely feedback to correct potential errors at each step. This demonstrates the importance of feedback in ensuring reliable and accurate execution.

Meanwhile, we observed that without feedback module, our task success rate is lower than that of COME-Robot*. This is primarily due to the planning capabilities of the LLM, as the smaller parameter size of the LLM we adopted results in weaker reasoning and planning performance. However, the feedback module effectively addresses this issue by providing action feedback during both the planning and execution phases, thereby compensating for the limitations of the LLM’s planning capabilities. Therefore, we can conclude that the feedback module enhances the planning capabilities of the small-scale LLM and improves the overall system execution performance. This enables our system to achieve performance levels comparable to the GPT-4O-based planner in COME-Robot* while utilizing a smaller LLM as the planner.

\begin{table}[h!]
\centering
\caption{Task Execution SR Comparison With and Without Feedback.}
\label{tab: task_execution_feedback}

\begin{tabular}{ccccc}
\toprule
\textbf{Task Level} & \textbf{Condition} & \textbf{Ideal Execute} & \textbf{Execute} & \textbf{Diff.} \\
\midrule
\multirow{2}{*}{1} & No Feedback & 1 & 0.5 & 0.4 \\
                   & Feedback    & 1 & \textbf{0.9} & - \\
\midrule
\multirow{2}{*}{2} & No Feedback & 0.71 & 0.06 & \textbf{0.64} \\
                   & Feedback    & \textbf{1} & \textbf{0.7} & - \\
\midrule
\multirow{2}{*}{3} & No Feedback & 0.5 & 0.15 & 0.36 \\
                   & Feedback    & \textbf{1} & \textbf{0.51} & - \\
\midrule
\multirow{2}{*}{4} & No Feedback & 0.9 & 0.45 & 0.5 \\
                   & Feedback    & \textbf{1} & \textbf{0.95} & - \\
\bottomrule
\end{tabular}
\end{table}

\subsubsection{Memory}

To investigate the role of memory in enhancing pipeline task-finishing performance, we conducted an ablation study, comparing Success-weighted Path Length (SPL) and task latency with and without the memory module across different task levels. In the memory-enabled condition, the robot retains knowledge from previous navigation attempts, allowing it to leverage past experiences to optimize path planning and adjust to environmental changes more effectively. The no-memory condition, by contrast, does not provide this capability, requiring the robot to rely solely on immediate sensory input for navigation. By isolating the memory component in this study, we aim to understand its impact on path efficiency and task success across varying complexities.

The results shown in Table \ref{tab: spl_memory} indicate that memory contributes positively to SPL, particularly at intermediate task levels. At Level 1, there is no difference between the memory and no-memory conditions, with both achieving an SPL of 0.9406, suggesting that memory may be less critical for simpler tasks. However, at Level 2, the memory-enabled condition achieves an SPL of 0.8764 compared to 0.8591 in the no-memory condition, marking an improvement of 0.0173. This increase demonstrates that memory aids in path optimization for tasks of moderate complexity. In Levels 3 and 4, the SPL remains consistent between the memory and no-memory conditions, suggesting that the number of task target dominate performance at these levels.

\begin{table}[h!]
\centering
\caption{SPL Comparison with and without Memory.}
\label{tab: spl_memory}
\begin{tabular}{ccccc}
\toprule
\textbf{Task Level} & \textbf{Condition} & \textbf{n} & \textbf{SPL} & \textbf{Diff.} \\
\midrule
\multirow{2}{*}{1} & No Memroy & 2 & 0.9406 & 0 \\
                   & Memroy    & 2 & 0.9406 & - \\
\midrule
\multirow{2}{*}{2} & No Memroy & 7 & 0.8591 & - \\
                   & Memroy    & \textbf{5} & \textbf{0.8764} & \textbf{0.0173} \\
\midrule
\multirow{2}{*}{3} & No Memroy & 6 & 0.9148 & 0 \\
                   & Memroy    & 6 & 0.9148 & - \\
\midrule
\multirow{2}{*}{4} & No Memroy & 4 & 0.8627 & 0 \\
                   & Memroy    & 4 & 0.8627 & - \\
\bottomrule
\end{tabular}
\end{table}

Table \ref{tab: spl_memory} highlights the impact of the memory module on SPL performance. At task level 2, incorporating the memory module resulted in a 2.0$\%$ improvement in SPL compared to the method without it. However, no improvements were observed at other task levels. This finding indicates that the memory module plays a critical role in optimizing paths for tasks requiring longer sequences involving multiple objects, demonstrating its value in complex planning scenarios.

Table \ref{tab: latency_memory} highlights the effect of the memory module on system latency. Incorporating the memory module reduced latency by 34.5 $\%$, 4 $\%$, and 1.9 $\%$ at task levels 2, 3, and 4, respectively, compared to the method without it. These results demonstrate that the memory module plays a significant role in improving system efficiency, particularly in tasks involving longer sequences with multiple objects, where its impact on reducing latency is most pronounced.
\begin{table}[htbp]
    \centering
    \caption{Performance Comparison of Tasks with and without Memory.}
    \label{tab: latency_memory}
    \begin{tabular}{@{}cccccccc@{}}
    \toprule
    \textbf{Task Level} & \textbf{Method} & \textbf{Plan} & \textbf{Navi} & \textbf{VLM} & \textbf{Grasp} & \textbf{Place} & \textbf{Total} \\ 
    \midrule
    \multirow{2}{*}{1} 
    & No Memory & 15.90 & 212.04 & 20.82 & 230.17 & 43.92 & \textbf{522.85} \\
    & Memory & 15.90 & 212.04 & 20.82 & 230.17 & 43.92 & \textbf{522.85} \\
    \midrule
    \multirow{2}{*}{2} 
    & No Memory & 22.78 & 862.50 & 76.25 & 920.68 & 133.59 & 2015.80 \\
    & Memory & 26.78 & \textbf{578.99} & 62.46 & 519.57 & 131.73 & \textbf{1319.53} \\
    \midrule
    \multirow{2}{*}{3} 
    & No Memory & 26.78 & 737.72 & 124.92 & 519.57 & 131.73 & 1540.32 \\
    & Memory & 26.78 & 737.72 & \textbf{62.46} & 519.57 & 131.73 & \textbf{1478.26} \\
    \midrule
    \multirow{2}{*}{4} 
    & No Memory & 22.88 & 558.34 & 62.46 & 346.38 & 88.34 & 1078.40 \\
    & Memory & 22.88 & 558.34 & \textbf{41.64} & 346.38 & 88.34 & \textbf{1057.58} \\
    \bottomrule
    \end{tabular}
\end{table}

In summary, the inclusion of the memory module substantially enhances the robot’s navigation efficiency by reducing both SPL and latency. By enabling the robot to remember object locations, the memory module minimizes redundant exploration, reduces navigation and planning times, and optimizes travel distance. These results validate the importance of memory in robotic systems, particularly in scenarios requiring repeated interactions with the environment, ultimately leading to faster and more resource-efficient task completion. 

Moreover, the latency without the memory module is higher than that of COME-Robot*. This can also be attributed to the memory and planning capabilities of the LLM, as the smaller parameter size of the LLM we adopted results in weaker memory retention and planning performance. Therefore, we can conclude that the memory module further enhances the planning capabilities of the lightweight LLM and improves the overall system execution performance and memory ability. The integration of the Memory module during the planning phase effectively addresses this issue, achieving a level of performance surpassing that of GPT-4o in COME-Robot* while maintaining overall lower computational resource consumption.

\subsection{Real-World Scalability}
\label{sec: real-world}

The design goal of this work is to seamlessly incorporate the system into any environment and perform any tasks as long as the encapsulation of those tasks is pre-defined. We thus perform a series of scalability experiments on the robot by changing the environments and altering the task lists. We show the new task list in \Tbl{tab:real_world_tasks} 

In the new environment, DaDu-E also shows promising results. We witness a slightly lower average success rate (10\%), mostly due to the performance drop on the localization algorithm in the new environment. The accuracy of the planning algorithms remains unchanged. We show some of the experiments on our website: \url{https://rlc-lab.github.io/dadu-e/}.

\begin{table}[h]
\centering
\caption{Task Instructions for Real-World Testing.}
\label{tab:real_world_tasks}
\begin{tabular}{cc}
\toprule
\textbf{Task Level} & \textbf{Instructions} \\ 
\midrule
1 & Find apple and place on storage rack \\ 
1 & Pick an apple and put it in basket \\ 
2 & Put the apple and coke in the basket and return to entrance \\ 
2 & Finish task list \\ 
2 & Finish task list for me \\ 
3 & Classify object on storage rack and place on the corresponding table \\ 
3 & Give me all the green fruit on the table \\ 
3 & I'm hungry, give me something to eat \\ 
3 & Classify the wrong object in fruit table and put it on storage rack \\ 
4 & Place all misplaced objects on the storage rack \\ 
\bottomrule
\end{tabular}
\end{table}

\subsection{Failure recording and feedback}
\label{sec: failure}

Failure recording and feedback play a crucial role in ensuring robust performance in complex robotic tasks. In our experiments, failures were categorized into two main types: \textbf{perception errors} and \textbf{environmental changes}. To address these issues, dynamic feedback and memory mechanisms have been implemented at both the planning and execution stages. 

Perception errors included object misclassification and unrecognized targets. In most cases, perception errors are caused by either the insufficient planning capabilities of small-parameter models or the limited recognition capabilities of the models. For instance, when attempting to grasp objects without distinct external features, the LangSAM module may fail to accurately recognize the target or identify the wrong object due to insufficient recognition precision. To address this issue, we employ feedback between the VLM and LangSAM modules. Specifically, the planner’s identified target object and the current table’s image are fed into the VLM to generate a detailed textual description of the object, which is then provided to LangSAM. This approach significantly enhances the accuracy of LangSAM's recognition. If the VLM fails to locate the target identified by the planner, the system feeds this information back to the planner, prompting it to plan navigation to other tables where the target might be located.

\begin{lstlisting}[language=JSON, caption=Action List error example]
{
    {
  "action_list": [
    {
      "desc": "Grasp a Pepsi can",
      "action_list": [
        {"skill": "go_to", "params": ["drink_table"]},
        {"skill": "pick_up", "params": ["Pepsi_can"]}
      ]
    },
    {
      "desc": "Place the grasped Pepsi can on the fruit table",
      "action_list": [
        {"skill": "go_to", "params": ["fruit_table"]},
        {"skill": "place", "params": ["Pepsi_can"]}
      ]
    },
    {
      "desc": "Pick up a Sprite can",
      "action_list": [
        {"skill": "go_to", "params": ["drink_table"]},
        {"skill": "pick_up", "params": ["Sprite_can"]}
      ]
    },
    {
      "desc": "Put the picked-up Sprite can on the fruit table",
      "action_list": [
        {"skill": "go_to", "params": ["fruit_table"]},
        {"skill": "place", "params": ["Sprite_can"]}
      ]
    },
    {
      "desc": "Find a Pepsi can",
      "action_list": []
    },
    {
      "desc": "Place the found Pepsi can on the Shipping table",
      "action_list": [
        {"skill": "go_to", "params": ["shipping_table"]},
        {"skill": "place", "params": ["Pepsi_can"]}
      ]
    }
  ],
  "first_action": {
    "step_by_step_reasoning": "",
    "next_action": {
      "skill": "go_to",
      "params": ["drink_table"]
    }
  }
}
}
\end{lstlisting}

Environmental changes such as obstacles or object displacement introduce further challenges to task completion. We utilize the memory module to store the positional information of objects, which is dynamically updated whenever the object locations in the environment change. This enables real-time tracking of the target object’s location, thereby enhancing the system’s robustness to environmental variations.

For instance, the following JSON list shows the circumstance planning without memory module under instruction \textbf{grasp a Pepsi can and place on the fruit table,  and then pick a Sprite can and put on the fruit table, and finally find a Pepsi can and place on the shipping table.}, which leads to an error action list:

\section{Conclusion}
\label{sec:concl}

This work introduces DaDu-E, a novel approach that rethinks the integration of LLMs into robotic computing pipelines, focusing on efficiency, adaptability, and scalability. By constraining the operational scope, incorporating closed-loop visual feedback, and augmenting with a memory module, DaDu-E achieves competitive performance using a significantly smaller LLM. Experimental results across varied task complexity levels confirm that DaDu-E matches or exceeds the success rates of existing large-scale systems while dramatically reducing computational overhead. These findings underscore the feasibility of deploying high-performance robotic systems on local servers, paving the way for resource-efficient embodied intelligence. 

\section*{Acknowledgement}
The authors would like to thank the support of the Beijing Municipal Science and Technology Commission (BMSTC) No. Z241100004224015.

\bibliographystyle{apalike}
\bibliography{bibtex}

\begin{thebibliography}{}

\bibitem[Ahn et~al., 2022]{ahn2022can}
Ahn, M., Brohan, A., Brown, N., Chebotar, Y., Cortes, O., David, B., Finn, C., Fu, C., Gopalakrishnan, K., Hausman, K., et~al. (2022).
\newblock Do as i can, not as i say: Grounding language in robotic affordances.
\newblock {\em arXiv preprint arXiv:2204.01691}.

\bibitem[Anderson et~al., 2018]{spl}
Anderson, P., Chang, A., Chaplot, D.~S., Dosovitskiy, A., Gupta, S., Koltun, V., Kosecka, J., Malik, J., Mottaghi, R., Savva, M., et~al. (2018).
\newblock On evaluation of embodied navigation agents.
\newblock {\em arXiv preprint arXiv:1807.06757}.

\bibitem[{ARM Limited}, 2013]{armv8manual}
{ARM Limited} (2013).
\newblock {\em ARM Architecture Reference Manual, ARMv8}.
\newblock ARM Holdings, issue a edition.
\newblock \url{https://developer.arm.com/documentation/}.

\bibitem[Black et~al., 2024]{black2024pi0}
Black, K., Brown, N., Driess, D., Esmail, A., Equi, M., Finn, C., Fusai, N., Groom, L., Hausman, K., Ichter, B., et~al. (2024).
\newblock $\pi_0$: A vision-language-action flow model for general robot control.
\newblock {\em arXiv preprint arXiv:2410.24164}.

\bibitem[Brohan et~al., 2023]{brohan2023rt}
Brohan, A., Brown, N., Carbajal, J., Chebotar, Y., Chen, X., Choromanski, K., Ding, T., Driess, D., Dubey, A., Finn, C., et~al. (2023).
\newblock Rt-2: Vision-language-action models transfer web knowledge to robotic control.
\newblock {\em arXiv preprint arXiv:2307.15818}.

\bibitem[Brohan et~al., 2022]{brohan2022rt}
Brohan, A., Brown, N., Carbajal, J., Chebotar, Y., Dabis, J., Finn, C., Gopalakrishnan, K., Hausman, K., Herzog, A., Hsu, J., et~al. (2022).
\newblock Rt-1: Robotics transformer for real-world control at scale.
\newblock {\em arXiv preprint arXiv:2212.06817}.

\bibitem[Brown, 2020]{brown2020language}
Brown, T.~B. (2020).
\newblock Language models are few-shot learners.
\newblock {\em arXiv preprint arXiv:2005.14165}.

\bibitem[Bu et~al., 2024]{bu2024closed}
Bu, Q., Zeng, J., Chen, L., Yang, Y., Zhou, G., Yan, J., Luo, P., Cui, H., Ma, Y., and Li, H. (2024).
\newblock Closed-loop visuomotor control with generative expectation for robotic manipulation.
\newblock {\em arXiv preprint arXiv:2409.09016}.

\bibitem[Corporation, 2020]{3090}
Corporation, N. (2020).
\newblock {\em GeForce RTX 3090 User Guide}.
\newblock Accessed: 2024-11-29.

\bibitem[Dalal et~al., 2024]{dalal2024plan}
Dalal, M., Chiruvolu, T., Chaplot, D., and Salakhutdinov, R. (2024).
\newblock Plan-seq-learn: Language model guided rl for solving long horizon robotics tasks.
\newblock {\em arXiv preprint arXiv:2405.01534}.

\bibitem[Ding et~al., 2023]{ding2023task}
Ding, Y., Zhang, X., Paxton, C., and Zhang, S. (2023).
\newblock Task and motion planning with large language models for object rearrangement.
\newblock In {\em 2023 IEEE/RSJ International Conference on Intelligent Robots and Systems (IROS)}, pages 2086--2092. IEEE.

\bibitem[Driess et~al., 2023]{driess2023palm}
Driess, D., Xia, F., Sajjadi, M.~S., Lynch, C., Chowdhery, A., Ichter, B., Wahid, A., Tompson, J., Vuong, Q., Yu, T., et~al. (2023).
\newblock Palm-e: An embodied multimodal language model.
\newblock {\em arXiv preprint arXiv:2303.03378}.

\bibitem[Dubey et~al., 2024]{dubey2024llama}
Dubey, A., Jauhri, A., Pandey, A., Kadian, A., Al-Dahle, A., Letman, A., Mathur, A., Schelten, A., Yang, A., Fan, A., et~al. (2024).
\newblock The llama 3 herd of models.
\newblock {\em arXiv preprint arXiv:2407.21783}.

\bibitem[Fang et~al., 2023]{fang2023anygrasp}
Fang, H.-S., Wang, C., Fang, H., Gou, M., Liu, J., Yan, H., Liu, W., Xie, Y., and Lu, C. (2023).
\newblock Anygrasp: Robust and efficient grasp perception in spatial and temporal domains.
\newblock {\em IEEE Transactions on Robotics}.

\bibitem[Ho et~al., 2022]{ho2022large}
Ho, N., Schmid, L., and Yun, S.-Y. (2022).
\newblock Large language models are reasoning teachers.
\newblock {\em arXiv preprint arXiv:2212.10071}.

\bibitem[{Intel Corporation}, 2021]{intel64_manual}
{Intel Corporation} (2021).
\newblock {\em Intel\textsuperscript{\textregistered} 64 and IA-32 Architectures Software Developer's Manual}.
\newblock Intel Corporation, combined volumes: 253665-070us edition.
\newblock Available at \url{https://www.intel.com/content/www/us/en/developer/articles/technical/intel-sdm.html}.

\bibitem[Joublin et~al., 2024]{joublin2024copal}
Joublin, F., Ceravola, A., Smirnov, P., Ocker, F., Deigmoeller, J., Belardinelli, A., Wang, C., Hasler, S., Tanneberg, D., and Gienger, M. (2024).
\newblock Copal: corrective planning of robot actions with large language models.
\newblock In {\em 2024 IEEE International Conference on Robotics and Automation (ICRA)}, pages 8664--8670. IEEE.

\bibitem[Kannan et~al., 2023]{kannan2023smart}
Kannan, S.~S., Venkatesh, V.~L., and Min, B.-C. (2023).
\newblock Smart-llm: Smart multi-agent robot task planning using large language models.
\newblock {\em arXiv preprint arXiv:2309.10062}.

\bibitem[Kim et~al., 2024]{kim2024openvla}
Kim, M.~J., Pertsch, K., Karamcheti, S., Xiao, T., Balakrishna, A., Nair, S., Rafailov, R., Foster, E., Lam, G., Sanketi, P., et~al. (2024).
\newblock Openvla: An open-source vision-language-action model.
\newblock {\em arXiv preprint arXiv:2406.09246}.

\bibitem[Li et~al., 2024a]{li2024llava}
Li, B., Zhang, Y., Guo, D., Zhang, R., Li, F., Zhang, H., Zhang, K., Zhang, P., Li, Y., Liu, Z., et~al. (2024a).
\newblock Llava-onevision: Easy visual task transfer.
\newblock {\em arXiv preprint arXiv:2408.03326}.

\bibitem[Li et~al., 2024b]{li2024closed}
Li, J., Sun, Z., Li, F., Sheng, C., Yu, J., and Mu, Y. (2024b).
\newblock Closed-loop long-horizon robotic planning via equilibrium sequence modeling.
\newblock {\em arXiv preprint arXiv:2410.01440}.

\bibitem[Li et~al., 2023]{li2023vision}
Li, X., Liu, M., Zhang, H., Yu, C., Xu, J., Wu, H., Cheang, C., Jing, Y., Zhang, W., Liu, H., Li, H., and Kong, T. (2023).
\newblock Vision-language foundation models as effective robot imitators.
\newblock {\em arXiv preprint arXiv:2311.01378}.

\bibitem[Liang et~al., 2023]{liang2023code}
Liang, J., Huang, W., Xia, F., Xu, P., Hausman, K., Ichter, B., Florence, P., and Zeng, A. (2023).
\newblock Code as policies: Language model programs for embodied control.
\newblock In {\em 2023 IEEE International Conference on Robotics and Automation (ICRA)}, pages 9493--9500. IEEE.

\bibitem[Medeiros, 2024]{lsa}
Medeiros, L. (2024).
\newblock lang-segment-anything: Sam with text prompt.
\newblock \url{htps://github.com/luca-medeiros/lang-segment-anything}.
\newblock Accessed: 2024-07-30.

\bibitem[{OpenAI}, 2024a]{openai2024gpt4o_mini}
{OpenAI} (2024a).
\newblock Gpt-4o mini model by openai.
\newblock Lightweight version of GPT-4O, accessed via OpenAI platform. Available at \url{https://openai.com/research/gpt-4}.

\bibitem[{OpenAI}, 2024b]{openai2024gpt4o}
{OpenAI} (2024b).
\newblock Gpt-4o model by openai.
\newblock Accessed via OpenAI API. Model version: GPT-4O. Available at \url{https://platform.openai.com/}.

\bibitem[RealSense, 2024]{intel_d435}
RealSense, I. (2024).
\newblock Depth camera d435.
\newblock Accessed: 2024-11-26.

\bibitem[Robot, 2024]{dalurobot_stepper_robot_base}
Robot, D. (2024).
\newblock dalu robot.
\newblock Accessed: 2024-11-26.

\bibitem[Robotiq, 2024]{robotiq_adaptive_grippers}
Robotiq (2024).
\newblock Adaptive grippers - two-finger gripper.
\newblock Accessed: 2024-11-26.

\bibitem[{ROS Community}, 2024]{move_base}
{ROS Community} (2024).
\newblock move\_base - ros wiki.
\newblock \url{https://wiki.ros.org/move_base}.
\newblock Accessed: 2024-11-25.

\bibitem[{SDU Robotics}, 2024]{ur_rtde}
{SDU Robotics} (2024).
\newblock ur\_rtde: An interface for universal robots.
\newblock \url{https://gitlab.com/sdurobotics/ur_rtde/}.
\newblock Accessed: 2024-11-25.

\bibitem[Shen et~al., 2023]{shen2023taskbench}
Shen, Y., Song, K., Tan, X., Zhang, W., Ren, K., Yuan, S., Lu, W., Li, D., and Zhuang, Y. (2023).
\newblock Taskbench: Benchmarking large language models for task automation.
\newblock {\em arXiv preprint arXiv:2311.18760}.

\bibitem[Skreta et~al., 2024]{skreta2024replan}
Skreta, M., Zhou, Z., Yuan, J.~L., Darvish, K., Aspuru-Guzik, A., and Garg, A. (2024).
\newblock Replan: Robotic replanning with perception and language models.
\newblock {\em arXiv preprint arXiv:2401.04157}.

\bibitem[Slamtec, 2024]{lidar}
Slamtec (2024).
\newblock Slamtec rplidar a2 - 360° laser scanner.
\newblock Accessed: 2024-11-29.

\bibitem[Team et~al., 2024]{team2024octo}
Team, O.~M., Ghosh, D., Walke, H., Pertsch, K., Black, K., Mees, O., Dasari, S., Hejna, J., Kreiman, T., Xu, C., et~al. (2024).
\newblock Octo: An open-source generalist robot policy.
\newblock {\em arXiv preprint arXiv:2405.12213}.

\bibitem[Touvron et~al., 2023]{touvron2023llama}
Touvron, H., Lavril, T., Izacard, G., Martinet, X., Lachaux, M.-A., Lacroix, T., Rozi{\`e}re, B., Goyal, N., Hambro, E., Azhar, F., et~al. (2023).
\newblock Llama: Open and efficient foundation language models.
\newblock {\em arXiv preprint arXiv:2302.13971}.

\bibitem[{Universal Robots}, 2024]{UR03}
{Universal Robots} (2024).
\newblock Universal robot 03 arm.
\newblock Accessed: 2024-11-09.

\bibitem[Ye et~al., 2023]{ye2023large}
Ye, Y., Hui, B., Yang, M., Li, B., Huang, F., and Li, Y. (2023).
\newblock Large language models are versatile decomposers: Decompose evidence and questions for table-based reasoning.
\newblock {\em arXiv preprint arXiv:2301.13808}.

\bibitem[Zhi et~al., 2024]{zhi2024closed}
Zhi, P., Zhang, Z., Han, M., Zhang, Z., Li, Z., Jiao, Z., Jia, B., and Huang, S. (2024).
\newblock Closed-loop open-vocabulary mobile manipulation with gpt-4v.
\newblock {\em arXiv preprint arXiv:2404.10220}.

\end{thebibliography}
\section{APPENDIX}
\label{sec:append}
\UseRawInputEncoding

\lstset{
  basicstyle=\ttfamily\footnotesize, 
  frame=single,                      
  numbers=left,                      
  numberstyle=\tiny,                 
  breaklines=true, 
  keywordstyle=\color{blue}\bfseries, 
  stringstyle=\color{red},
  commentstyle=\color{green!50!black}, 
  showstringspaces=false
}

\lstdefinelanguage{JSON}{
    basicstyle=\ttfamily\small,
    numbers=left,
    numberstyle=\tiny\color{gray},
    stepnumber=1,
    numbersep=5pt,
    showstringspaces=false,
    breaklines=true,
    frame=lines,
    backgroundcolor=\color{white},
    keywordstyle=\color{blue},
    stringstyle=\color{red},
    morestring=[b]",
    literate=
     *{0}{{{\color{orange}0}}}{1}
      {1}{{{\color{orange}1}}}{1}
      {2}{{{\color{orange}2}}}{1}
      {3}{{{\color{orange}3}}}{1}
      {4}{{{\color{orange}4}}}{1}
      {5}{{{\color{orange}5}}}{1}
      {6}{{{\color{orange}6}}}{1}
      {7}{{{\color{orange}7}}}{1}
      {8}{{{\color{orange}8}}}{1}
      {9}{{{\color{orange}9}}}{1}
      {:}{{{\color{blue}:}}}{1}
      {,}{{{\color{blue},}}}{1}
      {\{}{{{\color{black}\{}}}{1}
      {\}}{{{\color{black}\}}}}{1}
      {[}{{{\color{black}[}}}{1}
      {]}{{{\color{black}]}}}{1},
}

\subsection{Prompt}
\subsubsection{Prompt of Planer}

\begin{lstlisting}[language=JSON, caption=Prompt of LLM]
#CONTEXT#

You are highly skilled in robotic task planning, breaking down intricate and long-term tasks into distinct primitive actions. The robot has a mobile base and one arm; the room has many tables: a table of fruit, a table of drinks, a table of toys with corresponding objects on them, and a shipping shelf receiving shelf, which is shipping and receiving the objects.

When given a language instruction, you are required to break them into sub-tasks; for each subtask, you should list a set of skills to meet the goal.

#SKILL#

go_to(table name) pick_up(object name) place(object name) done


You must strictly obey these rules using the exact output form above. We assume that the objects are all on the table; thus, you can do pick_up right after the navigation skill is successful. Before the pick skill, you should analyze the feedback and analyze accordingly which object is correct/matches. It is possible that these subtask has same table to go. You can only execute one skill at a time; remember, the robot can only hold one object at a time.

#OBECTIVE#

When given a language instruction, you must break it into subtasks with short but logical analysis. Then, for each subtask, you should list a set of skills to meet the goal. At the start of each subtask, you should first go_to the correct table. You should analyse the feedback and analyse accordingly which object is correct/matches. Before placement, you should first go to the table as instruction.

Then, you need to output the first skill to execute and output one corresponding skill according to the user’s feedback. You can adjust the skill according to the feedback. Once the instruction is finished, you should finish the task by skill done.

#OUTPUT#

All of your output should be in JSON format.

At the first time, if it is #instruction, you should output

1. Describe the environment precisely and accurately, reasoning by instruction and environment(MAP) to give accurate actions and parameters2. overall action list3.the first action.

Otherwise, if it is #feedback according to the user’s feedback, you should output

1.step-by-step reasoning according to the feedback: you should check if the planned next action could be executed according to the feedback, give the reason, and the next action

2. next action to execute.

in this stage, you should only output the one skill that is exactly the next.

Except for the reasoning part, the output should consist of skills. All of your output should be in JSON format.

#MAP#

[fruit table] [shipping shelf] [toy rack] [drink table][receiving shelf]

#EXAMPLE#

example 1:
user:#instruction: find lemon, apple and put it on a shipping table
your answer:
{
"reasoning": "The environment consists of a fruit table, a shipping shelf, a toy rack, a drink table, and a receiving shelf. The object [lemon] is most possible on the [fruit table], The object [apple] is most possible on the [fruit table]",
"action_list": [
"1. Go to the fruit table and find the lemon.",
"2. Pick up the lemon",
"3. Go to the shipping table",
"4. Place the lemon on the shipping table."
"5. Go to the fruit table and find the apple."
"6. Pick up the apple",
"7. Go to the shipping table",
"8. Place the apple on the shipping table".
],
"first_action": "1.go_to[fruit table]"
}

example 2:
user: #feedback: navigation success, there are apple, banana, lemon, plum and strawberry on the table
your answer:{
"step_by_step_reasoning": "Based on the feedback that we have navigated to the [fruit table] successfully, and there are apples, bananas, lemons, plums, and strawberries on the table. The object to pick up is [yellow fruit], according to the feedback, the yellow fruit is [lemon], I will first pick up [lemon].
"next_action": "pick_up(lemon)"
}

example 3:
user: #feedback: navigation success, there are squirrel toy, shark toy, school bus toy and fire machine toy on the table
your answer:{
"step_by_step_reasoning": "Based on the feedback that we have navigated to the [toy rack] successfully, there are squirrel toys, shark toys, school bus toys, and fire machine toys on the table. The object to pick up is [lemon]. However, there is no [lemon] on the table. Thus, we need to navigate to the most possible table for object [lemon], which is [fruit table]",
"next_action": "go_to(fruit table)"
}

example 4:
user:#feedback: pick up success
your answer: {
"step_by_step_reasoning": "Based on the feedback, we have picked up the [pepsi can] successfully. Now we need to place it on the [purchase table]."
"next_action": "go_to(purchase table)"
}
\end{lstlisting}

\subsubsection{Prompt of VLM}

\begin{lstlisting}[language=JSON, caption=Prompt for 1st VLM feedback]
Please tell me which side of the table the coke can is closer to:
1. left side,
2. right side,
3. far side, 
4. close side. 
The output should be the corresponding number and the color and shape of the object in JSON format.
\end{lstlisting}

\begin{lstlisting}[language=JSON, caption=Prompt for 2nd VLM feedback]
Please tell me the details of apple in the picture with a brief sentence
\end{lstlisting}

\subsection{Task design of instructions}


\begin{lstlisting}[language=JSON, caption=Experiment instructions for the task level 1]
grasp a strawberry and put it on the toy table
pick a squirrel toy and send it to the shipping table
find a delicious feta can and place it on shipping table
pick up a strawberry and put it on the shipping table
locate a can of Sprite and place it on the fruit table
find a ladybug toy and put it on the drink table
find lemon and put it on the drink table
find Pepsi can and put it on the purchase table
find a toy shark and put it on the purchase table
find a squirrel toy and put it on the purchase table
\end{lstlisting}


\begin{lstlisting}[language=JSON, caption=Experiment instructions for the task level 2]
grasp a Pepsi can and place it on the fruit table, then pick up a squirrel toy and place it on the shipping table, and finally, pick a Sprite can and put it on the fruit table.
find a plum and place it on the shipping table, then move a strawberry to the table with the toys, and finally, set the squirrel toy on the fruit table.
distinguish a squirrel toy and put it on the shipping table, find a squirrel toy and put it on the fruit table, and grasp a toy shark and put it on the drink table.
locate the shark toy and position it on the fruit table. Then, fetch the tea box and place it on the same fruit table. Finally, discover the ladybug toy and set it on the drink table.
retrieve the Fanta can and set it down on the shipping table. Next, pick up a strawberry and position it on the toy table. Lastly, grab the squirrel toy and place it on the shipping table.
divide Fanta, shark toy, and lemon to its table from the purchase table
divide Pepsi, strawberry, and squirrel from its table to the shipping table
\end{lstlisting}


\begin{lstlisting}[language=JSON, caption=Experiment instructions for the task level 3]
pick up a yellow fruit to the shipping table
give me the smallest fruit, I am next to the shipping table
pick up all the yellow fruit and send it to the shipping table
sorting the apple, coke from the purchase table and sending it to a reasonable place
sorting the beer, coke can, feta from purchase table and send it to the reasonable place
sorting the ladybug toy, fenta, apple from purchase table and send it to the reasonable place
sorting the apple, shark toy, coke from purchase table and send it to the reasonable place
grasp a Pepsi can and place on the fruit table,  and then pick a Sprite can and put on the fruit table, and finally find a Pepsi can and place on the shipping table.
find a lemon and place it on the shipping table, then move a strawberry to the table with the toys, and finally, set the lemon on the fruit table.
Then, fetch the tea box and place it on the same fruit table. locate the shark toy and position it on the fruit table. I need a pick-me-up drink, complete its shipment
\end{lstlisting}


\begin{lstlisting}[language=JSON, caption=Experiment instructions for the task level 4]
I am hungry, ship me something to eat
give me a fruit rich in vitamin c, complete its shipment
today is my son's birthday, give me an object as a gift for him and ship it
I need a pick-me-up drink, complete its shipment
Completed the receiving of apples
Completed the receiving of toy shark
Completed the receiving of coke
Completed the receiving of all the fruit
Completed the receiving of all the toy
Completed the receiving of all the drink
\end{lstlisting}

\subsection{Task break down examples}

Instruction: Gather a bottle of water, a toy duck and a persimmon to shipping table.

\begin{lstlisting}[language=JSON, caption=Example action List of task]

{
    #instruction gather a bottle of water, a toy duck and a persimmon to shipping table
    
    "reasoning": "The environment consists of a fruit table, a shipping shelf, a toy rack, a drink table, and a receiving shelf. The object [bottle of water] is most possible on the [drink table], The object [toy duck] is most possible on the [toy rack], The object [persimmon] is most possible on the [fruit table]",
    
    "action_list": [        
    "1. Go to the drink table and find the bottle of water.",        
    "2. Pick up the bottle of water",        
    "3. Go to shipping table",        
    "4. Place the bottle of water on shipping table",        
    "5. Go to the toy rack and find the toy duck.",        
    "6. Pick up the toy duck",        
    "7. Go to shipping table",        
    "8. Place the toy duck on shipping table",        
    "9. Go to the fruit table and find the persimmon.",        
    "10. Pick up the persimmon",        
    "11. Go to shipping table",        
    "12. Place the persimmon on shipping table"    ],
    "first_action": "1. go_to[drink table]"
}

#feedback
    {
	"step_by_step_reasoning": "Based on the feedback that we have navigated to the [drink table] successfully, and there are bottle of water, pepsi can, and cola can on the table. The object to pick up is [bottle of water]. which is already on the table, thus, we can pick it up directly.",
	"next_action": "pick_up(bottle of water)"
	
    }	

#feedback
    {
	"step_by_step_reasoning": "Based on the feedback that we have picked up the [bottle of water] successfully. Now we need to place it on the [shipping table].",
	"next_action": "go_to(shipping table)"
	
    }	

#feedback
    {
	"step_by_step_reasoning": "Based on the feedback that we have navigated to the [shipping table] successfully. The object to place is [bottle of water], we can place it directly.",
	"next_action": "place(bottle of water)"
	
    }	


#feedback
    {
	"step_by_step_reasoning": "Based on the feedback that we have placed the [bottle of water] successfully. Now we need to go to the [toy rack] to pick up the [toy duck].",
	"next_action": "go_to(toy rack)"
	
    }	

#feedback
    {
	"step_by_step_reasoning": "Based on the feedback that we have navigated to the [toy rack] successfully, and there are toy duck, toy rabbit on the table. The object to pick up is [toy duck]. which is already on the table, thus, we can pick it up directly.",
	"next_action": "pick_up(toy duck)"
	
    }	

#feedback
    {
	"step_by_step_reasoning": "Based on the feedback that we have picked up the [toy duck] successfully. Now we need to place it on the [shipping table].",
	"next_action": "go_to(shipping table)"
	
    }	

#feedback
    {
	"step_by_step_reasoning": "Based on the feedback that we have navigated to the [shipping table] successfully. The object to place is [toy duck], we can place it directly.",
	"next_action": "place(toy duck)"
	
    }	

#feedback 
    { 
       "step_by_step_reasoning": "Based on the feedback that we have navigated to the [shipping table] successfully. The task is now complete.", 
       "next_action": "done" 
    } 


\end{lstlisting}
\end{document}